\documentclass{article}

\usepackage{microtype}
\usepackage{graphicx}
\usepackage{subfigure}
\usepackage[dvipsnames]{xcolor}
\usepackage{booktabs, tabularx, wrapfig, adjustbox, multirow} % for professional tables
\usepackage{amssymb}
\usepackage{footmisc}

\newcolumntype{L}[1]{>{\raggedright\arraybackslash}p{#1}}
\newcolumntype{C}[1]{>{\centering\arraybackslash}p{#1}}
\newcolumntype{R}[1]{>{\raggedleft\arraybackslash}p{#1}}

\usepackage{hyperref}

\usepackage[accepted]{icml2021}

\icmltitlerunning{Self-Damaging Contrastive Learning}

\begin{document}

\newcommand\Tianlong[1]{\textcolor{blue}{[TL: #1]}}
\newcommand\Ziyu[1]{\textcolor{red}{[#1]}}

\twocolumn[
\icmltitle{Self-Damaging Contrastive Learning}




\begin{icmlauthorlist}
\icmlauthor{Ziyu Jiang}{to}
\icmlauthor{Tianlong Chen}{goo}
\icmlauthor{Bobak Mortazavi}{to}
\icmlauthor{Zhangyang Wang}{goo}
\end{icmlauthorlist}

\icmlaffiliation{to}{Texas A\&M University}
\icmlaffiliation{goo}{University of Texas at Austin}

\icmlcorrespondingauthor{Zhangyang Wang}{atlaswang@utexas.edu}
% \icmlcorrespondingauthor{Eee Pppp}{ep@eden.co.uk}

\icmlkeywords{Machine Learning, ICML}

\vskip 0.3in
]



\printAffiliationsAndNotice{}  % leave blank if no need to mention equal contribution

\begin{abstract}
% The recent advance achieved of Contrastive Learning accelerate the pace for deploying unsupervised training on real applications, motivating the study of its property under widely exist long tail distribution. In this paper, we explore the balanceness of Contrastive Learning under multiple challenging long tail settings and for the first time find that Contrastive Learning still suffers from the imbalance data distribution. Further, we address the imbalancess by proposing a framework termed as Self-Damaging Contrastive Learning Framework leveraging the property of Pruning Identified Exemplars (PIEs) \cite{hooker2020compressed}. Our extensive experiments across multiple datasets show that this method can improve both balanceness and accuracy.

The recent breakthrough achieved by contrastive learning accelerates the pace for deploying unsupervised training on real-world data applications. However, unlabeled data in reality is commonly imbalanced and shows a long-tail distribution, and it is unclear how robustly the latest contrastive learning methods could perform in the practical scenario. This paper proposes to explicitly tackle this challenge, via a principled framework called \textit{Self-Damaging Contrastive Learning} (\textbf{SDCLR}), to automatically balance the representation learning without knowing the classes. Our \textit{main inspiration} is drawn from the recent finding that  deep models have difficult-to-memorize samples, and those may be exposed through network pruning \cite{hooker2020compressed}. It is further natural to hypothesize that long-tail samples are also tougher for the model to learn well due to insufficient examples. 
%In addition to creating strong contrastive views by input data augmentation, 
Hence, the \textit{key innovation} in SDCLR is to create a dynamic \textit{self-competitor} model to contrast with the target model, which is a pruned version of the latter. During training, contrasting the two models will lead to adaptive online mining of the most easily forgotten samples for the current target model, and implicitly emphasize them more in the contrastive loss. Extensive experiments across multiple datasets and imbalance settings show that SDCLR significantly improves not only overall accuracies but also balancedness, in terms of linear evaluation on the full-shot and few-shot settings. Our code is available at \url{https://github.com/VITA-Group/SDCLR}. %Remarkably, in CIFAR100-LT, compared to baseline, the proposed method can improve the linear evaluation accuracy by 7.15\% while improving the balancedness.
\vspace{-1em}
\end{abstract}

\vspace{-1em}
\section{Introduction}

% Paragraph 1. Introduce the success of general SSL learning; 
\vspace{-0.2em}
\subsection{Background and Research Gaps}
\vspace{-0.3em}
Contrastive learning \cite{chen2020simple,he2020momentum,grill2020bootstrap,jiang2020robust,you2020graph} recently prevails for deep neural networks (DNNs) to learn powerful visual representations from unlabeled data. The state-of-the-art contrastive learning frameworks consistently benefit from using bigger models and training on more task-agnostic unlabeled data \cite{chen2020big}. The predominant promise implied by those successes is to leverage contrastive learning techniques to pre-train strong and transferable representations from internet-scale sources of unlabeled data. However, going from the controlled benchmark data to uncontrolled real-world data will run into several gaps. For example, most natural image and language data exhibit a Zipf long-tail distribution where various feature attributes have very different occurrence frequencies  \cite{zhu2014capturing,feldman2020does}. Broadly speaking, such imbalance is not only limited to the standard single-label classification with majority versus minority class \cite{liu2019large}, but also can extend to multi-label problems along many attribute dimensions \cite{sarafianos2018deep}. That naturally questions whether contrastive learning can still generalize well in those long-tail scenarios.

\begin{figure*}[t]
%\vspace{-0.5em}
  \centering
  \includegraphics[trim = 0mm 0mm 0mm 0mm, clip,scale=0.53]{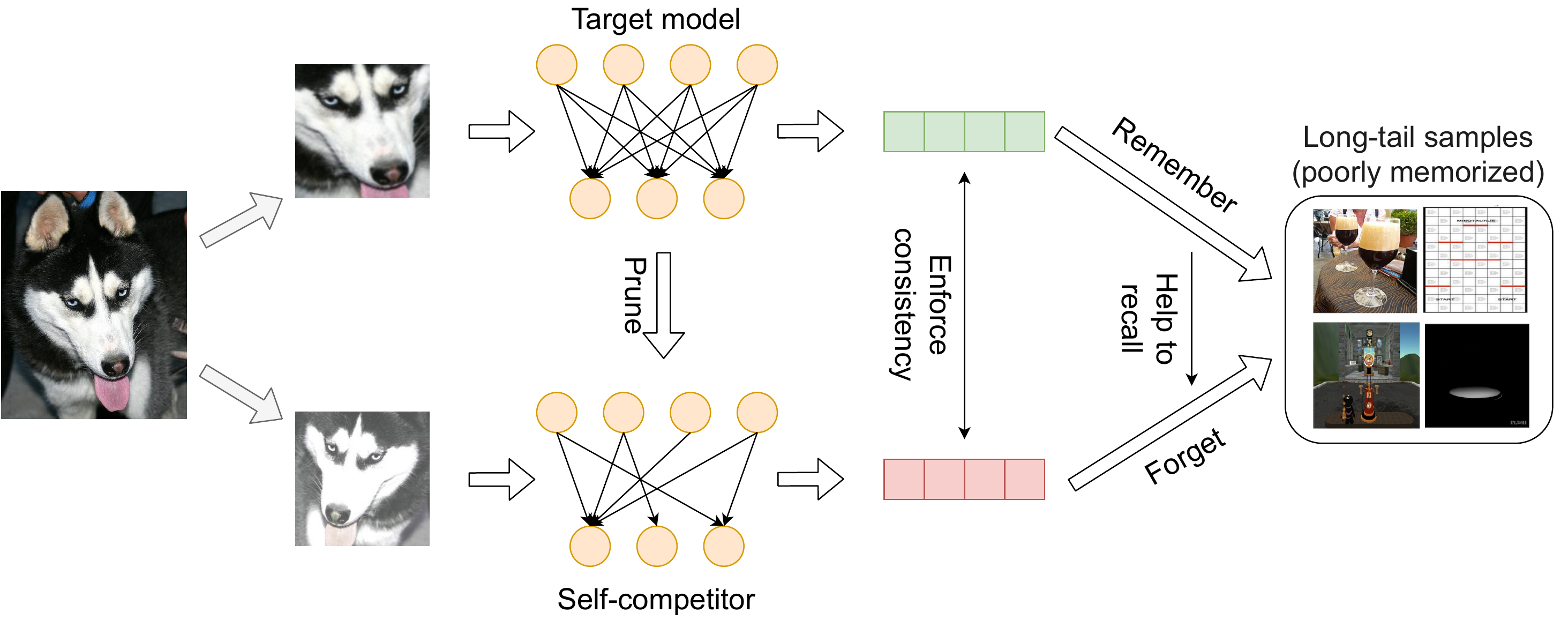}
  %\vspace{-0.5em}
    \caption{The overview of the proposed SDCLR framework. Built on top of simCLR pipeline \cite{chen2020simple} by default, the uniqueness of SDCLR lies in its two different network branches: one is the target model to be trained, and the other ``self-competitor" model that is pruned from the former online. The two branches share weights for their non-pruned parameters. Either branch has its independent batch normalization layers. Since the self-competitor is always obtained and updated from the latest target model, the two branches will co-evolve during training. Their contrasting will implicitly give more weights on long-tail samples.}
    %one is dense network and the other is its sparse version. Note that two networks share the same weight in the non-pruned part except BN layers, for which we employ independent parameters for the two networks.
 %   \vspace{-1em}
  \label{fig:selfdamaging_structure}
\end{figure*}
%is one of the most successful Self-supervised learning methods in deep neural networks (DNNs). It not only can learn strong visual representation without expensive human labels, but it is also proven to be helpful when serving as initialization for downstream tasks \cite{chen2020simple,he2020momentum,grill2020bootstrap}.

% Paragraph SSL is also proved to be helpful for imbalance setting. 
% Paragraph 2. Recent literary \cite{kang2021exploring} points out that SSL learning can generate balanced feature spaces even it is trained on extremely long tail datasets. However, this contradicts the observation in ... . This paradox motivates us to further check the balanceness of SSL learning.

We are \textit{not} the first to ask this important question. Earlier works ~\cite{yang2020rethinking,kang2021exploring} pointed out that when the data is imbalanced by class, contrastive learning can learn more balanced feature space than its supervised counterpart. Despite those preliminary successes, we find that the state-of-the-art contrastive learning methods remain certain vulnerability to the long-tailed data (even indeed improving over vanilla supervised learning), after digging into more experiments and imbalance settings (see Sec 4). Such vulnerability is reflected on the linear separability of pre-trained features (the instance-rich classes has much more separable features than instance-scarce classes), and affects downstream tuning or transfer performance. To conquer this challenge further, the main hurdle lies in the absence of class information; therefore, existing approaches for supervised learning, such as re-sampling the data distribution \cite{shen2016relay,mahajan2018exploring}  or re-balancing the loss for each class \cite{khan2017cost,cui2019class,cao2019learning}, cannot be straightforwardly made to work here. 

\vspace{-0.5em}
\subsection{Rationale and Contributions}
\vspace{-0.2em}
Our overall goal is to find a bold push to extend the loss re-balancing and cost-sensitive learning ideas \cite{khan2017cost,cui2019class,cao2019learning} into an unsupervised setting. The \textbf{initial hypothesis} arises from the recent observations that DNNs tend to prioritize learning simple patterns \cite{zhang2016understanding, arpit2017closer,liu2020early,yao2020searching,han2020sigua,xia2021robust}. More precisely, the DNN optimization is content-aware, taking advantage of patterns shared by more training examples, and therefore inclined towards memorizing the majority samples. Since long-tail samples are underrepresented in the training set, they will tend to be poorly memorized, or more ``easily forgotten" by the model - a characteristic that one can potentially leverage to spot long-tail samples from unlabeled data in a model-aware yet class-agnostic way. 

However, it is in general tedious, if ever feasible, to measure how well each individual training sample is memorized in a given DNN \cite{carlini2019secret}. One \textbf{blessing} comes from the recent empirical finding \cite{hooker2020compressed} in the context of image classification. The authors observed that, network pruning, which usually removes the smallest-magnitude weights in a trained DNN, does not affect all learned classes or samples equally. Rather, it tends to disproportionally hamper the DNN memorization and generalization on the long-tailed and most difficult images from the training set. In other words, long-tail images are not ``memorized well" and may be easily ``forgotten" by pruning the model, making network pruning a practical tool to spot the samples not yet well learned or represented by the DNN.

Inspired by the aforementioned, we present a principled framework called \textit{Self-Damaging Contrastive Learning} (\textbf{SDCLR}), to automatically balance the representation learning without knowing the classes. The workflow of SDCLR is illustrated in Fig. \ref{fig:selfdamaging_structure}. In addition to creating strong contrastive views by input \textit{data augmentation}, SDCLR introduces another new level of contrasting via``\textit{model augmentation}, by perturbing the target model's structure and/or current weights. In particular, the \textbf{key innovation} in SDCLR is to create a \textit{dynamic self-competitor} model by pruning the target model online, and contrast the pruned model's features with the target model's.  Based on the observation \cite{hooker2020compressed} that pruning impairs model ability to predict accurately on rare and atypical instances, those samples in practice will also have the largest prediction differences before then pruned and non-pruned models. That effectively boosts their weights in the contrastive loss and leads to \textbf{implicit loss re-balancing}. Moreover, since the self-competitor is always obtained from the updated target model, the two models will \textbf{co-evolve}, which allows the target model to spot diverse memorization failures at different training stages and to progressively learn more balanced representations. Below we outline our main contributions:\vspace{-0.6em}
\begin{itemize}
    \item Seeing that unsupervised contrastive learning is \textit{not immune} to the imbalance data distribution, we design a Self-Damaging Contrastive Learning (SDCLR) framework to address this new challenge.\vspace{-0.3em} 
    \item SDCLR innovates to leverage the latest advances in understanding DNN memorization. By creating and updating a self-competitor online by pruning the target model during training, SDCLR provides an adaptive online mining process to always focus on the most easily forgotten (long tailed) samples throughout training.\vspace{-0.3em}
    %We design the Self-Damaging Contrastive Learning (SDCLR) framework to address imbalance without labels. It can online adaptively mine the most easily forgotten samples by contrasting with its pruned network. While the most forgotten samples heavily overlap with long-tail samples. \cite{hooker2020compressed}, this network can implicity weight more on the tail-classes, which leads to higher balancedness.
    \item Extensive experiments across multiple datasets and imbalance settings show that SDCLR can significantly improve not only the balancedness of the learned representation.\vspace{-0.5em}
  %  framework can improve the balanceness of contrastive learning features (by linear evaluation accuracy), but it also boosts the practical downstream few-shot tuning performance. effectively improve both balancesness and accuracy of Contrastive Learning when considering the linear evaluation on the full dataset and the few-shot fine-tuning performance. 
\end{itemize}

%Instead of finding some other strong external competitors, the \textbf{key step and merit} in our approach is to compress the target model using network pruning techniques \cite{han2015learning, he2019filter, li2016pruning,liu2017learning}, to construct strong ``self-competitors" from the same target model. 

% Tackling the imbalanceness problem under unsupervised setting becomes much more challenging. The traditional methods like re-sampling and re-weighting techniques can not be directly used given the lack of the labels. One can estimate the label via clustering, but it is very likely to introduce undesired noise. Fortunately, a recent work \cite{hooker2020compressed} reveals that networks tend to forget a small subset of hard and atypical samples, which are termed as Pruning Identified Exemplars (PIEs). The most challenging samples in the long tail settings are the sample in the tail classes. Thus, enhancing the feature of PIEs could help to enhance the feature of the most hard tail samples. 

%Our extensive experiments on multiple datasets shows that this model can not only improve the balanceness of contrastive learning features, but it also boosts the overall performance. In conclusion, our contributions are:

\section{Related works}
\vspace{-0.2em}
\textbf{Data Imbalance and Self-supervised Learning:} Classical long-tail recognition mainly amplify the impact of tail-class samples, by re-sampling or re-weighting \cite{cao2019learning, cui2019class, chawla2002smote}. However, those hinge on label information and are not directly applicable to unsupervised representation learning. Recently, \cite{kang2019decoupling,zhang2019balance} demonstrate the learning of feature extractor and classifier head can be decoupled. 
%and it is crucial to learn well-balanced feature extraction for long-tailed recognition, e.g., through a class-balanced sampling. 
That suggests the promise of pre-training a feature extractor. Since it is independent of any later task-driven fine-tuning stage, such strategy is compatible with any existing imbalance handling techniques in supervised learning.

%The ubiquitousness of long-tail problem in the real world; The development for long-tail classification.To resolve the imbalanceness under unsupervised setting, one straightforward method is to explicitly enlarge the gradient of tail samples by re-sampling or re-weighting \cite{cao2019learning, cui2019class, chawla2002smote}. However this can not be directly achieved due to the lack of the labels under unsupervised setting. One can estimate the label via clustering, but it is very likely to introduce undesired noise.

Inspired by so, latest works start to explore the benefits of a balanced feature space from self-supervised pre-training for generalization. 
\cite{yang2020rethinking} presented the first study to utilize self-supervision for overcoming the intrinsic label bias. They observe that simply plugging in self-supervised pre-training, e.g., rotation prediction \cite{gidaris2018unsupervised} or MoCo \cite{he2020momentum}, would outperform their corresponding end-to-end baselines for long tailed classification. Also given more unlabeled data, the labels can be more effectively leveraged in a semi-supervised manner for accurate and debiased classification. 
reduce label bias in a semi-supervised manner. Another positive result was reported in a (concurrent) piece of work~\cite{kang2021exploring}. The authors pointed out that when the data is imbalanced by class, contrastive learning can learn more balanced feature space than its supervised counterpart.% where the representations present similar linear separability w.r.t. all classes. 

\textbf{Pruning as Compression and Beyond: } DNNs can be compressed of excessive capacity \cite{lecun1990optimal} at surprisingly little sacrifice of test set accuracy, and various pruning techniques \cite{han2015deep,li2016pruning,liu2017learning}  have been popular and effective for that goal. Recently, some works have notably reflected on pruning beyond just an ad-hoc compression tool, exploring its deeper connection with DNN memorization/generalization. \cite{frankle2018lottery} showed that there exist highly sparse ``critical subnetworks" from the full DNNs, that can be trained in isolation from scratch to reaching the latter's same performance. That critical subnetwork could be identified by iterative unstructured pruning \cite{frankle2019lottery}. 

The most relevant work to us is \cite{hooker2020compressed}. For a trained image classifier, pruning it has a non-uniform impact: a fraction of classes, which usually belong to the ambiguous/difficult classes, or the long-tail of less frequent instances. are disproportionately impacted by the introduction of sparsity. That provides novel insights and means to exposing a trained model's weakness in generalization. For example, \cite{wang2021troubleshooting} leveraged this idea to construct an ensemble of self-competitors from one dense model, to troubleshoot an image quality model in the wild.

\textbf{Contrasting Different Models:} The high-level idea of SDCLR, i.e., contrasting two similar competitor models and weighing more on their most disagreed samples, can trace a long history back to the selective sampling framework \cite{atlas1990training}. One most fundamental algorithm is the seminal \textit{Query By Committee} (QBC) \cite{seung1992query,gilad2005query}. During learning, QBC maintains a space of classifiers that are consistent on predicting all previous labeled samples. At a new unlabeled example, QBC selects two random hypotheses from the space and only queries for the label of the new example if the two disagree. In comparison, our focused problem is in the different realm of unsupervised representation learning.

Spotting two models' disagreement for troubleshooting either is also an established idea \cite{popper1963science}. 
%The underlying scientific philosophy is rooted in the classical notion of ``model falsification as model comparison” by \cite{popper1963science}. 
That concept has an interesting link to the popular technique of differential testing \cite{mckeeman1998differential} in software engineering. The idea has also been applied to model comparison and error-spotting in computational vision \cite{wang2008maximum} and image classification \cite{wang2019going}. However, none of those methods has considered to construct a self-competitor from a target model. They also work in a supervised active learning setting rather than unsupervised.

Lastly, co-teaching \cite{han2018co,yu2019does} performs sample selection in noisy label learning by using two DNNs, each trained on a different subset of examples that have a small training loss for the other network. Its limitation is that the examples that are selected tend to be easier, which may slow down learning \cite{chang2017active} and hinder generalization to more difficult data \cite{song2019selfie}. On the opposite, our method is designed to focus on the difficult-to-learn samples in the long tail.

%Compression disproportionately impacts model performance on the underrepresented long-tail of the data distribution. PIEs over-index on atypical or noisy images that are far more challenging for both humans and algorithms to classify

% We also provide some insight into the role of these additional parameters, which appear necessary to encode a useful representation of the long-tail low frequencydata points.
\vspace{-0.5em}
\section{Method}
% Paragraph 1. Organization introduction
% In this section, we would first introduce the motivation and implementation of the proposed Self-Damaging Contrastive Learning. Afterwards, we  
\vspace{-0.3em}
% Paragraph 2. Notation introduction
\subsection{Preliminaries}
\vspace{-0.2em}
\textbf{Contrastive Learning.}
% Paragraph 1.  Introduce the general framework of SSL.
Contrastive learning learns visual representation via enforcing similarity of the positive pairs $(v_{i}, v_{i}^{+})$ and enlarging the distance of negative pairs $(v_{i}, v_{i}^{-})$. Formally, the loss is defined as
\begin{equation}
    \mathcal{L}_{\mathrm{CL}}=\frac{1}{N} \sum_{i=1}^{N}-\log \frac{s \left(v_{i} ,v_{i}^{+}, \tau\right)}{s \left(v_{i}, v_{i}^{+}, \tau\right)+\sum_{v_{i}^{-} \in V^{-}} s \left(v_{i}, v_{i}^{-}, \tau\right)}
\end{equation}
where  $s \left(v_{i} ,v_{i}^{+}, \tau\right)$ indicates the similarity between positive pairs while $s \left(v_{i} ,v_{i}^{-}, \tau\right)$ is the similarity between negative pairs. $\tau$ represents the temperature hyper-parameter. The negative samples $v_{i}^{-}$ are sampled from negative distribution $V^{-}$. The similarity metric is typically defined as
\begin{equation}
    s \left(v_{i}, v_{i}^{+}, \tau\right) = \exp \left(v_{i} \cdot v_{i}^{+} / \tau\right)
\end{equation}

SimCLR \cite{chen2020simple} is one of the state-of-the-art contrastive learning frameworks. For an input image, SimCLR would augment it twice with two different augmentations, and then process them with two branches  that share the same architecture and weights. Two different versions of the same image are set as positive pairs, and the negative image is sampled from the rest images in the same batch.

\textbf{Pruning Identified Exemplars.} \cite{hooker2020compressed} systematically investigates the model output changes introduced by pruning and finds that certain examples are particularly sensitive to sparsity. These images most impacted after pruning are termed as \textit{Pruning Identified Exemplars} (\textbf{PIEs}), representing the difficult-to-memorize samples in training. Moreover, the authors also demonstrate that PIEs often show up at the long-tail of a distribution.

We extend \cite{hooker2020compressed}'s PIE hypothesis from supervised classification to the unsupervised setting for the first time. Moreover, instead of pruning a trained model and expose its PIEs once, we are now integrating pruning into the training process as an online step. With PIEs dynamically generated by pruning a target model under training, we expect them to expose different long-tail examples during training, as the model continues to be trained. Our experiments show that PIEs answer well to those new challenges.

%are more challenging to be recognized by model or human. Moreover, they also demonstrate that PIEs often show up at the long-tail of a distribution. The hypothesis of this phenomena is that compression impairs model ability to predict accurately on rare and atypical instances. 

%We take inspiration from \cite{hooker2020compressed}, and improve their methodology to troubleshoot BIQA models, by identifying and leveraging quality-discriminable images between pruned and non-pruned methods. 

%To avoid any confusion, we stress that \textbf{we are NOT using pruning for any model compression or efficiency purpose}.

%\textbf{key assumption}: sara's work in unsupervised setting to expose long-tail examples, and can work online during training (rather than for a trained model)

% Paragraph 1. Introduce the PIE samples
% Pruning Identified Exemplars (PIEs) are defined as the samples that one network and its compressed version disagree on. In other words, they are subset of images that is impacted the most by the sparsity. PIEs are found to over-index on most challenging samples like atypical or noisy images. 

\vspace{-0.2em}
\subsection{Self-Damaging Contrastive Learning}
\vspace{-0.2em}
% In addition to creating strong contrastive views by input augmentation, e.g. creating  the key step and merit in our
% approach is to compress the target model using network
% pruning techniques [15, 16, 24, 27], to construct strong “selfcompetitors” from the same target model. The critical underlying rationale takes root in the recent finding [17] in the
% context of image classification, The authors observed that,
% network pruning, which usually removes smallest-magnitude
% weights in a trained network, does not affect all learned
% classes or samples equally. Rather, it tends to disproportionally hamper the network memorization and generalization
% on the long-tailed and most difficult images from the training
% distribution. In other words, those images are not “memorized well” by the current model, may be easily “forgotten”
% by pruning the model. Therefore, network pruning can effectively spot the samples not yet well learned or represented,
% hence exposing the current trained model’s weakness

\textbf{Observation: Contrastive learning is NOT immune to imbalance.} Long-tail distribution fails many supervised approaches build on balanced benchmarks \cite{kang2019decoupling}. Even contrastive learning does not rely on class labels, it still learns the transformation invariances in a data-driven manner, and will be affected by dataset bias \cite{purushwalkam2020demystifying}. Particularly for long-tail data, one would naturally hypothesize that the instance-rich head classes may dominate the invariance learning procedure and leaves the tail classes under-learned. 

The concurrent work \cite{kang2021exploring} signaled that using the contrastive loss can obtain a balanced representation space that has similar separability (and downstream classification performance) for all the classes, backed by experiments on ImageNet-LT \cite{liu2019large} and iNaturalist \cite{van2018inaturalist}. We independently reproduced and validated their experimental findings. However, we have to point out that it was pre-mature to conclude ``contrastive learning is immune to imbalance". 

To see that, we present additional experiments in Section \ref{CLbiased2LT}. While that conclusion might hold for a moderate level of imbalance as presented in current benchmarks, we have constructed a few heavily imbalanced data settings, in which cases contrastive learning will become unable to produce balanced features. In those case, the linear separability of learned representation can differ a lot between head and tail classes. We suggest that our observations complement those in \cite{yang2020rethinking,kang2021exploring}, that while (vanilla) contrastive learning can to some extent alleviate the imbalance issue in representation learning, \textit{it does not possess full immunity} and calls for further boosts.

\textbf{Our SDCLR Framework. } 
% Our experiments in Section. \ref{exp} show that the imbalanceness still exists for the representation trained with Contrastive Learning. To address the imbalanceness, we propose a framework called Self-Damaging Contrastive Learning (SDCLR) based on the aforementioned finding of Pruning Identified Exemplars (PIEs) as they are more likely to be found at the long-tail classes.
% Paragraph 2. The design of pruning for contrastive learning framework
% In our settings, the most challenging samples exist in the tail classes and more complex classes. Therefore, we hypothesize that PIEs are mainly composed by these samples. This means we can enhance the feature of hard and tails samples via enhancing the features of PIE samples. Therefore, we design the self-damaging contrastive learning that can online mining the PIE samples and enhance their feature during the learning progress. 
Figure \ref{fig:selfdamaging_structure} overviews the high-level workflow of the proposed SDCLR framework. By default, SDCLR is built on top of the simCLR pipeline \cite{chen2020simple}, and follows its most important components such as data augmentations and non-linear projection head. The \underline{main difference} between simCLR and SDCLR lies in that, simCLR feeds the two augmented images into the same target network backbone (via weight sharing); while SDCLR creates a ``self-competitor" by pruning the target model online, and lets the two different branches take the two augmented images to contrast their features.

Specifically, at each iteration we will have a dense branch $N_1$, and a sparse branch $N^p_2$ by pruning $N_1$, using the simplest magnitude-based pruning as described in \cite{han2015deep}, following the practice of \cite{hooker2020compressed}. Ideally, the pruning mask of $N^p_2$ could be updated per iteration after the model weights are updated. In practice, since the backbone is a large DNN and its weights will not change much for a single iteration or two, we set the pruning mask to be lazy-updated at the beginning of every epoch, to save computational overheads; all iterations in the same epoch then adopt the same mask\footnote{We also tried to update pruning masks more frequently, and did not find observable performance boosts.}. Since the self-competitor is always obtained and updated from the latest target model, the two branches will co-evolve during training.  

We sample and apply two different augmentation chains to the input image $I$, creating two different versions [$\hat{I}_1$, $\hat{I}_2$]. They are encoded by [$N_1$, $N^p_2$], and their output features [$f_1$, $f^p_2$] are fed into the nonlinear projection heads to enforce similarity be under the NT-Xent loss \cite{chen2020simple}. Ideally, if the sample is well-memorized by $N_1$, pruning $N_1$ will not ``forget" it -- thus little extra perturbation will be caused and the contrasting is roughly the same as in the original simCLR. Otherwise, for rare and atypical instances, SDCLR will amplify the prediction differences between then pruned and non-pruned models -- hence those samples' weights be will implicitly increased in the overall loss. 

When updating the two branches, note that [$N_1$, $N^p_2$] will share the same weights in the non-pruned part, and $N_1$ will independently update the remaining part (corresponding to weights pruned to zero in $N^p_2$). Yet, we empirically discover that it helps to let either branch have its independent batch normalization layers, as the features in dense and sparse may show different statistics \cite{yu2018slimmable}.

\vspace{-0.2em}
\subsection{More Discussions on SDCLR}
\vspace{-0.2em}
\textbf{SDCLR can work with more contrastive learning frameworks.} We focus on implementing SDCLR on top of simCLR for proving the concept. However, our idea is rather plug-and-play and can be applied with almost every other contrastive learning framework adopting the the two-branch design \cite{he2020momentum,lecun1990optimal,grill2020bootstrap}. We will explore combining SDCLR idea with them as our immediate future work.

%Though we are focusing on SimCLR in this paper, our main idea is to mine the PIEs with one dense and one sparse network. As the most Contrastive Learning frameworks adopt the two-branch design \cite{he2020momentum,lecun1990optimal,grill2020bootstrap}, our method can plug-and-play for other Contrastive Learning frameworks.

\textbf{Pruning is NOT for model efficiency in SDCLR. } To avoid possible confusion, we stress that \textit{we are NOT using pruning for any model efficiency purpose}. In our framework, pruning would be better described as ``selective brain damage”. It is mainly used for effectively spotting samples not yet well memorized and learned by the current model. However, as will be shown in Section \ref{ablationExps}, the pruned branch can have a ``side bonus", that sparsity itself can be an effective regularizer that improves few-shot tuning.

\textbf{SDCLR benefits beyond standard class imbalance.} We also want to draw awareness that SDCLR can be extended seamlessly beyond the standard single-class label imbalance case. Since SDCLR relies on no label information at all, it is readily applicable to handling various more complicated forms of imbalance in real data, such as the multi-label attribute imbalance \cite{sarafianos2018deep,yun2021re}. 

Moreover, even in artificially class-balanced datasets such as ImageNet, there hide more inherent forms of ``imbalance", such as the class-level difficulty variations or instance-level feature distributions  \cite{bilal2017convolutional,beyer2020we}. Our future work would explore SDCLR in those more subtle imbalanced learning scenarios in the real world.

%The long-tail data distribution is not the only factor why the representation becomes imbalance. Another important factor is the varying difficulty of classes. As the challenging samples are also included in PIEs, our method can improve the performance of hard classes and thus boost the performance on balanced datasets.

\vspace{-0.5em}
\section{Experiments}
\label{exp}
\vspace{-0.2em}
\subsection{Datasets and Training Settings}
\vspace{-0.3em}
Our experiments are based on three popular imbalanced datasets at varying scales: long-tail CIFAR-10, long-tail CIFAR-100 and ImageNet-LT. Besides, to further stretch out contrastive learning's imbalance handling ability, we also consider a more realistic and more challenging benchmark long-tail ImageNet-100 as well as another long tail ImageNet with a different exponential sampling rule. The long-tail ImageNet-100 contains less classes, which decreases the number of classes that looks similar and thus can be more vulnerable to imbalance.
% we construct a new, more realistic and more challenging benchmark, \textbf{ImageNet-SC-LT} (SC stands for super class).

%We also construct a more challenging large scale dataset called ImageNet-SC-LT. ImageNet-SC-LT is imbalance in the superclass level, which leads to larger visual discrepancy between head and tail classes. Based on the best of our knowledge, this is the first time that such dataset is proposed. 

\textbf{Long-tail CIFAR10/CIFAR100:} The original CIFAR-10/ CIFAR-100 datasets consist of 60000 32$\times$32 images in 10/100 classes. Long tail CIFAR-10/ CIFAR-100 (CIFAR10-LT / CIFAR100-LT) were first introduced in \cite{cui2019class} by sampling long tail subsets from the original datasets. Its imbalance factor is defined as the class size of the largest class divided by the smallest class. We by default consider a challenging setting with the imbalance factor set as 100.
To alleviate randomness, all experiments are conducted with five different long tail sub-samplings.

%To remove the influence brought by the randomness of sub-sampling, Our experiments are conducted on five different long tail sub-samplings.

\textbf{ImageNet-LT:} ImageNet-LT is a widely used benchmark introduced in \cite{liu2019large}. The sample number of each class is determined by a Pareto distribution with the power value $\alpha=6$. The resultant dataset contains 115.8K images, with the sample number per class ranging from 1280 to 5.  

\textbf{ImageNet-LT-exp:} Another long tail distribution of ImageNet we considered is given by an exponential function \cite{cui2019class}, where the imbalanced factor set as 256 to ensure the the minor class scale is the same as ImageNet-LT. The resultant dataset contains 229.7K images in total. \footnote{\label{note1}Refer to our code for details of ImageNet-LT-exp and ImageNet-100-LT.}

\textbf{Long tail ImageNet-100:} In many fields such as medical, material, and geography, constructing an ImageNet scale dataset is expensive and even impossible. Therefore, it is also worth considering
% testing and improving the balancedness for 
a dataset with a small scale and large resolution. We thus sample a new long tail dataset called ImageNet-100-LT from ImageNet-100 \cite{tian2019contrastive}. The sample number of each class is determined by a down-sampled (from 1000 classes to 100 classes) Pareto distribution used for ImageNet-LT. The dataset contains 12.21K images, with the sample number per class ranging from 1280 to 5\footref{note1}.

To evaluate the influence brought by long tail distribution, for each long tail subset, we would sample a balanced subset from the corresponding full dataset with the same total size as the long tail one to disentangle the influences of long tail and sample size. 

\begin{table*}
\centering
\caption{Comparing the \textit{linear separability performance} for models learned on balanced subset $D_b$ and long-tail subset $D_i$ of CIFAR10 and CIFAR100. \textit{Many}, \textit{Medium} and \textit{Few} are split based on class distribution of the corresponding $D_i$.}
\vspace{3mm}
    \label{tab:showImbalance}
    \begin{adjustbox}{max width=\textwidth}
    \begin{tabular}{@{}L{1.5cm}C{1.5cm}C{2.5cm}C{2.5cm}C{2.5cm}C{2.5cm}@{}}
    \toprule
    Dataset  & Subset &  \textit{Many} & \textit{Medium} & \textit{Few} & All \\ \midrule

    \multirow{2}{*}{CIFAR10}  & $D_b$  & 82.93 $\pm$ 2.71 & 81.53 $\pm$ 5.13 & 77.49 $\pm$ 5.09 & 80.88 $\pm$ 0.16 \\
                              & $D_i$  & 78.18 $\pm$ 4.18 & 76.23 $\pm$ 5.33 & 71.37 $\pm$ 7.07 & 75.55 $\pm$ 0.66 \\\midrule 
    \multirow{2}{*}{CIFAR100} & $D_b$  & 46.83 $\pm$ 2.31 & 46.92 $\pm$ 1.82 & 46.32 $\pm$ 1.22 & 46.69 $\pm$ 0.63 \\  
                              & $D_i$  & 50.10 $\pm$ 1.70 & 47.78 $\pm$ 1.46 & 43.36 $\pm$ 1.64 & 47.11 $\pm$ 0.34 \\
   
    \bottomrule
    \end{tabular}
    \end{adjustbox}
    \vspace{-3mm}
\end{table*}

\begin{table*}
\centering
\caption{Comparing the \textit{few-shot performance} for models learned on balanced subset $D_b$ and long-tail subset $D_i$ of CIFAR10 and CIFAR100. \textit{Many}, \textit{Medium} and \textit{Few} are split according to class distribution of the corresponding $D_i$.}
\vspace{1mm}
    \label{tab:showImbalanceFewshot}
    \begin{adjustbox}{max width=\textwidth}
    \begin{tabular}{@{}L{1.5cm}C{1.5cm}C{2.5cm}C{2.5cm}C{2.5cm}C{2.5cm}@{}}
    \toprule
    Dataset  & Subset &  \textit{Many} & \textit{Medium} & \textit{Few} & All \\ \midrule
    \multirow{2}{*}{CIFAR10}  & $D_b$  & 77.14 $\pm$ 4.64 & 74.25 $\pm$ 6.54 & 71.47 $\pm$ 7.55 & 74.57 $\pm$ 0.65 \\  
                              & $D_i$  & 76.07 $\pm$ 3.88 & 67.97 $\pm$ 5.84 & 54.21 $\pm$ 10.24& 67.08 $\pm$ 2.15 \\ \midrule 
    \multirow{2}{*}{CIFAR100} & $D_b$  & 25.48 $\pm$ 1.74 & 25.16 $\pm$ 3.07  & 24.01 $\pm$ 1.23 & 24.89 $\pm$ 0.99 \\
                              & $D_i$  & 30.72 $\pm$ 2.01 & 21.93 $\pm$ 2.61 & 15.99 $\pm$ 1.51 & 22.96 $\pm$ 0.43 \\
    \bottomrule
    \end{tabular}
    \end{adjustbox}
\vspace{-3mm}
\end{table*}

For all pre-training, we follow the SimCLR recipe \cite{chen2020simple} including its augmentations, projection head structures. The default pruning ratio is 90\% for CIFAR and 30\% for ImageNet. We adopt Resnet-18 \cite{he2016deep} for small datasets (CIFAR10/CIFAR100), and Resnet-50 for larger datasets (ImageNet-LT/ImageNet-100-LT), respectively. More details on hyperparameters can be found in the supplementary.

\subsection{How to Measure Representation Balancedness}
\label{sec:balanceMeasure}
% Paragraph 1. The evaluation methods: balanceness definition; 

The balancedness of a feature space can be reflected by the linear separability w.r.t. all classes. To measure the linear separability, we identically follow \cite{kang2021exploring} to employ a three-step protocol: \textit{i)} learn the visual representation $f_v$ on the training dataset  with $\mathcal{L}_{CL}$. \textit{ii)} training a linear classifier layer $L$ on the top of $f_v$ with a labeled balanced dataset (by default, the full dataset where the imbalanced subset is sampled from). \textit{iii)} evaluating the accuracy of the linear classifier $L$ on the testing set. 
%In this way, this protocol quantifies the linear separability of $f_v$ to the accuracy of the linear layer $L$. 
Hereinafter, we define such accuracy measure as the \textit{\textbf{linear separability performance}}.

To better understand the influence of the balancedness for down-stream tasks, we consider the important practical application of few-shot learning \cite{chen2020big}. The only difference between measuring few-shot learning performance and measuring \textit{linear separability accuracy} lies in step \textit{ii)}: we use only 1\% samples of the full dataset from which the pre-training imbalanced dataset is sampled. Hereinafter, we define the accuracy measure with this protocol as the \textit{\textbf{few-shot performance}}.

We further divide each dataset to three disjoint groups in terms of the size of classes: \{\textit{Many}, \textit{Medium}, \textit{Few}\}. In subsets of CIFAR10/CIFAR100, \textit{Many} and \textit{Few} each include the largest and smallest $\frac{1}{3}$ classes, respectively. For instance in CIFAR-100: the classes with [500-106, 105-20, 19-5] samples belong to [Many (34 classes), Medium (33 classes), Few (33 classes)] categories, respectively. In subsets of ImageNet,  we follow OLTR \cite{liu2019large} to define \textit{Many} as classes each with over training 100 samples, \textit{Medium}  as classes each with 20-100 training samples and \textit{Few} as classes under 20 training samples. We report the average accuracy for each specified group, and also use the standard deviation (\textit{\textbf{Std}}) among the three groups' accuracies as another balancedness measure.

%use the group name to represent the average accuracy the specified group. We also use the Standard deviation (\textit{Std}) among average accuracy of the three groups to quantify the balancedness, by default we multiple \textit{Std} with 100 to align with the the accuracy ($\%$) scale.

\subsection{Contrastive Learning is NOT Immune to Imbalance}
\label{CLbiased2LT}

\begin{table*}[t]
\centering
\caption{
% Comparing the \textit{linear separability performance} and \textit{few-shot performance} for models learned on balanced subset $D_b$ and long-tail subset $D_i$ of ImageNet and ImageNet-100. \textit{Many}, \textit{Medium} and \textit{Few} are split according to class distribution of the corresponding $D_i$.
Comparing the \textit{linear separability performance} and \textit{few-shot performance} for models learned on balanced subset $D_b$ and long-tail subset $D_i$ of  ImageNet and ImageNet-100. We consider two long tail distributions for ImageNet: Pareto and Exp, which corresponds to ImageNet-LT and Imagenet-LT-exp, respectively. \textit{Many}, \textit{Medium} and \textit{Few} are split according to class distribution of the corresponding $D_i$.
}
\vspace{1mm}
    \label{tab:showImbalanceImagenet}
    \begin{adjustbox}{max width=\textwidth}
    \begin{tabular}{@{}L{2.5cm}C{1.8cm}C{1.4cm}C{1cm}C{1cm}C{1cm}C{1cm}C{1cm}C{1cm}C{1cm}C{1cm}@{}}
    \toprule
    \multirow{2}{*}{Dataset}      & \multirow{2}{*}{Long tail type}  & \multirow{2}{*}{Split type}    &  \multicolumn{4}{c}{\textit{linear separability}} &  \multicolumn{4}{c}{\textit{few-shot}} \\ \cmidrule(lr){4-7} \cmidrule(lr){8-11}
                                  &                           &      &  \textit{Many} & \textit{Medium} & \textit{Few}   & All & \textit{Many} & \textit{Medium} & \textit{Few} & All \\ 
    \midrule
    \multirow{2}{*}{ImageNet}     & \multirow{2}{*}{\textit{Pareto}} & $D_b$  & 58.03 & 56.02  & 56.71 & 56.89 & 29.26 & 26.97 & 27.82 & 27.97  \\  
                                  &  & $D_i$  & 58.56 & 55.71  & 56.66 & 56.93 & 31.36 & 26.21 & 27.21 & 28.33  \\ \midrule
    \multirow{2}{*}{ImageNet}     & \multirow{2}{*}{\textit{Exp}} & $D_b$  & 57.46 & 57.70 & 57.02 & 57.42 & 32.31 & 32.91 & 32.17 & 32.45 \\  
                                  &  & $D_i$  & 58.37 & 56.97 & 56.27 & 57.43 & 35.98 & 29.56 & 28.02 & 32.12  \\ \midrule
    \multirow{2}{*}{ImageNet-100} & \multirow{2}{*}{\textit{Pareto}}  & $D_b$  & 68.87 & 66.33  & 61.85 & 66.74 & 48.82 & 44.71 & 41.08 & 45.84  \\
                                  &  & $D_i$  & 69.54 & 63.71 & 59.69  & 65.46 & 48.36 & 39.00 & 35.23 & 42.16 \\
    \bottomrule
    \end{tabular}    
    \end{adjustbox}
\vspace{-4mm}
\end{table*}

We now investigate if contrastive learning is vulnerable to the long-tail distribution. In this section, we use $D_i$ to represent the long tail split of a dataset while $D_b$ denotes its balanced counterpart.
% with imbalance factor of 100 and test the \textit{linear separability performance} for models pre-trained on such dataset. For comparison, we also test \textit{linear separability performance} for models pre-trained on a balanced subset $D_b$ with the same number of samples as $D_i$. To alleviate the randomness, we repeat each experiment with different sampling for $K$ times. In practice, $K$ is set as 4 for CIFAR 100, and it's enlarged to 10 for CIFAR10 since the different splits in CIFAR10 always lead to larger variance.

\begin{table*}
\centering
\caption{Compare the proposed SDCLR  with SimCLR in terms of the \textit{linear separability performance}. $\uparrow$ means the metric the higher the better and $\downarrow$ means the metric is the lower the better.}
\vspace{3mm}
    \label{tab:pruneImproveImbalance}
    \begin{adjustbox}{max width=\textwidth}
    \begin{tabular}{@{}L{2.2cm}C{1.4cm}C{2.0cm}C{2.0cm}C{2.0cm}C{2.0cm}C{2.0cm}@{}}
    \toprule
    Dataset   & Framework &  \textit{Many} $\uparrow$ & \textit{Medium} $\uparrow$ & \textit{Few} $\uparrow$ & \textit{Std} $\downarrow$ & All $\uparrow$ \\ \midrule
    \multirow{2}{*}{CIFAR10-LT}      &   SimCLR   & 78.18 $\pm$ 4.18 & 76.23 $\pm$ 5.33 & 71.37 $\pm$ 7.07 &  5.13 $\pm$ 3.66 & 75.55 $\pm$ 0.66 \\
                                     &   SDCLR    & 86.44 $\pm$ 3.12 & 81.84 $\pm$ 4.78 & 76.23 $\pm$ 6.29 &  5.06 $\pm$ 3.91 & 82.00 $\pm$ 0.68 \\\midrule 
    \multirow{2}{*}{CIFAR100-LT}     &   SimCLR   & 50.10 $\pm$ 1.70 & 47.78 $\pm$ 1.46 & 43.36 $\pm$ 1.64 &  3.09 $\pm$ 0.85 & 47.11 $\pm$ 0.34 \\ 
                                     &   SDCLR    & 58.54 $\pm$ 0.82 & 55.70 $\pm$ 1.44 & 52.10 $\pm$ 1.72 &  2.86 $\pm$ 0.69 & 55.48 $\pm$ 0.62 \\\midrule 
    \multirow{2}{*}{ImageNet-100-LT} &   SimCLR   & 69.54            & 63.71            & 59.69            &    4.04          & 65.46            \\ 
                                     &   SDCLR    & 70.10            & 65.04            & 60.92            &    3.75          & 66.48            \\ 
    \bottomrule
    \end{tabular}
    \end{adjustbox}
\vspace{-2mm}
\end{table*}

\begin{table*}
\centering
\caption{Compare the proposed SDCLR  with SimCLR in terms of the \textit{few-shot performance}. $\uparrow$ means the metric the higher the better and $\downarrow$ means the metric is the lower the better.}
\vspace{3mm}
    \label{tab:pruneImproveImbalanceFewshot}
    \begin{adjustbox}{max width=\textwidth}
    \begin{tabular}{@{}L{2.5cm}C{1.4cm}C{2.0cm}C{2.0cm}C{2.0cm}C{2.0cm}C{2.0cm}@{}}
    \toprule
    Dataset & Framework & \textit{Many} $\uparrow$ & \textit{Medium} $\uparrow$ & \textit{Few} $\uparrow$ & \textit{Std} $\downarrow$ & All $\uparrow$ \\ \midrule
    \multirow{2}{*}{CIFAR10}         & SimCLR     & 76.07 $\pm$ 3.88 & 67.97 $\pm$ 5.84 & 54.21 $\pm$ 10.24& 9.80 $\pm$ 5.45 & 67.08 $\pm$ 2.15 \\  
                                     & SDCLR      & 76.57 $\pm$ 4.90 & 70.01 $\pm$ 7.88 & 62.79 $\pm$ 7.37 &  6.99 $\pm$ 5.20 & 70.47 $\pm$ 1.38 \\ \midrule
    \multirow{2}{*}{CIFAR100}        & SimCLR     & 30.72 $\pm$ 2.01 & 21.93 $\pm$ 2.61 & 15.99 $\pm$ 1.51 & 6.27 $\pm$ 1.20 & 22.96 $\pm$ 0.43 \\
                                     & SDCLR      & 29.72 $\pm$ 1.52 & 25.41 $\pm$ 1.91 & 20.55 $\pm$ 2.10 & 3.98 $\pm$ 0.98 & 25.27 $\pm$ 0.83 \\ \midrule
    \multirow{2}{*}{Imagenet-100-LT} &   SimCLR   & 48.36            & 39.00            & 35.23            &    5.52          & 42.16            \\ 
                                     &   SDCLR    & 48.31            & 39.17            & 36.46            &    5.07          & 42.38            \\ 
    \bottomrule
    \end{tabular}
    \end{adjustbox}
    % \vspace{-3mm}
\end{table*}

As shown in Table.~\ref{tab:showImbalance}, 
% for both CIFAR10 and CIFAR100, models pre-trained on $D_b$ show almost the same accuracy for three groups. However, when the pre-training subset switches from $D_b$ to $D_i$, models trained on both datasets show a similar phenomenon: the accuracy gradually drops from \textit{many} to \textit{few}. This indicates that the balancedness of contrastive learning is still fragile when trained over the long tail distributions.
for both CIFAR10 and CIFAR100, models pre-trained on $D_i$ show larger imbalancedness than that on $D_b$. For instance, in CIFAR100, while models pre-trained on $D_b$ show almost the same accuracy for three groups, the accuracy gradually drops from \textit{many} to \textit{few} when pre-training subset switches from $D_b$ to $D_i$. This indicates that the balancedness of contrastive learning is still fragile when trained over the long tail distributions.

% In CIFAR10, the accuracy of \textit{Many} on decrease by 4.29\% when switching from $D_b$ to $D_i$ while the accuracy of \textit{Medium} and \textit{Few} decrease by 8.96\% and 7.61\%, respectively. In CIFAR100, the effect of long-tail distribution is more obvious. While the model pre-trained on $D_b$ show almost the same accuracy on for three groups. The model pretrained on $D_i$ gradually drop from 50.22\% to 44.26\% as the sample number decrease.

We next explore if the imbalanced representation would influence the downstream few-shot learning applications. As shown in Table.~\ref{tab:showImbalanceFewshot}, in CIFAR10, the \textit{few shot performance} of \textit{Many} drops by 1.07\% when switching from $D_b$ to $D_i$ while that of \textit{Medium} and \textit{Few} decrease by 5.30\% and 6.12\%. In CIFAR100, when pre-training with $D_b$, the \textit{few-shot performance} on three groups are similar, and it would become imbalanced when the pre-training dataset switches from $D_b$ to $D_i$. In a word, the balancedness of few-shot performance is consistent with the representation balancedness. Moreover, the bias would become even more serious: The gap between \textit{Many} and \textit{Few} enlarge from 6.81\% to 21.86\% on CIFAR10 and from 6.65\% to 14.73\% on CIFAR100.

We further study if the imbalance can also influence large scale dataset like ImageNet in Table.~\ref{tab:showImbalanceImagenet}. For ImageNet-LT and Imagenet-LT-exp, while the imbalancedness of \textit{linear separability performance} shows weak, that problem becomes much more significant for \textit{few-shot performance}. Especially, for Imagenet-LT-exp, the \textit{few-shot performance} of \textit{Many} is 7.96\% higher than that of \textit{Few}. The intuition behind this is that the large volume of the balanced fine-tuning dataset could mitigate the influence of imbalancedness from the pre-trained model. When the scale decreases to 100 classes (ImageNet-100), the imbalancedness consistently exists and it can be reflected via both \textit{linear separability performance} and \textit{few-shot performance}.

% Exp. With protocol in \cite{kang2021exploring}. The SSL with long tail distribution in CIFAR10-LT/CIFAR100-LT/Imagenet-LT can show imbalanceness from another perspective. Pervious metric is not enough for measuring the imbalanceness because ...

\subsection{SDCLR Improves Both Accuracy and Balancedness on Long-tail Distribution}

We compare the proposed SDCLR with SimCLR \cite{chen2020simple} on the datasets that are most easily to be impacted by long tail distribution: CIFAR10-LT, CIFAR100-LT, and ImageNet-100-LT. 
%We first pre-train the model following SimCLR and SDCLR. Afterwards, we consider two settings to evaluate the performance of pre-training: 1. Fix the backbone and only fine-tuning the linear evaluation layer on the corresponding full dataset. 2. Free all the layers and fine-tuning on the balanced few-shot sub-sampling $S_f$.  The first one is for evaluating the learned representation while the second one is for evaluating the performance of the proposed method on the important application of Contrastive Learning.
As shown in Table.~\ref{tab:pruneImproveImbalance}, the proposed SDCLR leads to a significant \textit{linear separability performance} improvement of $\textbf{6.45}\%$ in CIFAR10-LT and $\textbf{8.37}\%$ in CIFAR100-LT. Meanwhile, SDCLR also improve the balancedness by reducing the \textit{Std} by $0.07\%$ in CIFAR10 and $0.23\%$ in CIFAR100. In Imagenet-100-LT, SDCLR achieve an improvement on \textit{linear separability performance} of $1.02\%$ while reducing the \textit{Std} by $0.29$.
% \Ziyu{Imagenet exps description to be added here.}
% \Ziyu{need new figure to visualize PIEs and show they belong to long-tail classes indeed}

On few-shot settings, as shown in Table.~\ref{tab:pruneImproveImbalanceFewshot}, the proposed SDCLR consistently improves the \textit{few-shot performance} by [$3.39\%$, $2.31\%$, $0.22\%$] while decreasing the \textit{Std} by [$2.81$, $2.29$, $0.45$] in [CIFAR10, CIFAR100, Imagenet-100-LT], respectively.

\subsection{SDCLR Helps Downstream Long Tail Tasks}
SDCLR is a pre-training approach that is fully compatible with almost any existing long-tail algorithm. To show that, on CIFAR-100-LT with the imbalance factor of 100, we use SDCLR as pre-training, to fine-tune a SOTA long-tail algorithm RIDE~\cite{wang2020long} on its top. With SDCLR pre-training, the overall accuracy can reach 50.56\%, super-passing the original RIDE result by 1.46\%. Using SimCLR pre-training for RIDE only spots 50.01\% accuracy. 

\subsection{SDCLR Improves Accuracy on Balanced Datasets}
\begin{table}
\centering
\caption{Compare the accuracy ($\%$) and Standard deviation (\textit{Std}) among classes in balanced CIFAR10/100. $\uparrow$ means the metric the higher the better; $\downarrow$ means the metric is the lower the better.}
\vspace{3mm}
    \label{tab:SDCLRimproveBalanceFullDataset}
    \begin{adjustbox}{max width=\linewidth}
    \begin{tabular}{@{}L{1.7cm}C{1.7cm}C{1.7cm}C{1.7cm}@{}}
    \toprule
    Datasets                     & Framework  & Accuracy $\uparrow$   &  Std $\downarrow$    \\ \midrule
    \multirow{2}{*}{CIFAR10}     & SimCLR     & 91.16      &  6.37   \\ 
                                 & SDCLR      & 91.55      &  5.37   \\ \midrule
    \multirow{2}{*}{CIFAR100}    & SimCLR     & 62.84      &  14.94  \\
                                 & SDCLR      & 66.32      &  14.82  \\
    % \multirow{2}{*}{ImageNet100} & SimCLR     &            &         \\
    %                              & SDCLR      &            &         \\
    \bottomrule
    \end{tabular}    
    \end{adjustbox}
\end{table}

Even balanced in sample numbers per class, existing datasets can still suffer from more hidden forms of ``imbalance", such as sampling bias, and different classes' difficulty/ambiguity levels, e.g., see~\cite{bilal2017convolutional,beyer2020we}. To evaluate whether the proposed SDCLR can address such imbalancedness, we further run the proposed framework on balanced datasets: The full dataset of CIFAR10 and CIFAR100. We compare SDCLR with SimCLR following standard linear evaluation protocol \cite{chen2020simple} (On the same dataset, it first pre-trains the backbone, and then finetunes one linear layer on the top of the output features).

The results are shown in Table~\ref{tab:SDCLRimproveBalanceFullDataset}. Note here the $Std$ denotes the standard deviation of classes as we are studying the imbalance caused by the varying difficulty of classes. The proposed SDCLR can boost the linear evaluation accuracy by [$0.39\%$, $3.48\%$] while reducing the Std by [$1.0$, $0.16$] in [CIFAR10, CIFAR100], respectively, proving that the proposed method can also help to improve the balancedness even in the balanced datasets. 

% \subsection{Balanceness improvement can help more}
% Paragraph 1. With protocol in \cite{kang2021exploring}. The pruning method can improve the balanceness from all perspectives for CIFAR10-LT/CIFAR100-LT/Imagenet-LT.

% Show that the pruning method can help more compared to previous SSL methods for CIFAR10-LT/CIFAR100-LT/Imagenet-LT benchmarks.

% \subsection{Why suppressing PIE samples helps}

\subsection{SDCLR Mines More Samples from The Tail}

\begin{figure}[t]
  \centering
   \vspace{-0.5em}
  \includegraphics[trim = 0mm 0mm 0mm 0mm, clip,scale=0.645]{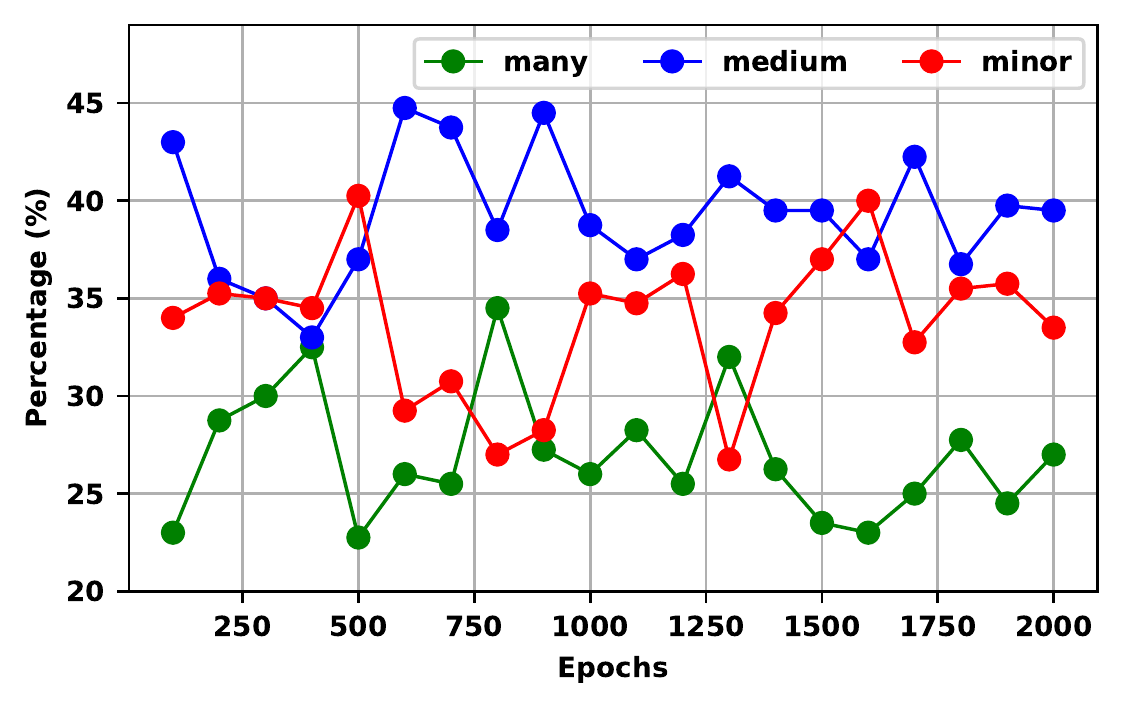}
  \vspace{-1em}
    \caption{Pre-training on imbalance splits of CIFAR100, The percentage of \textit{many} ($\textcolor{OliveGreen}{\bullet}$), \textit{medium} ($\textcolor{blue}{\bullet}$) and \textit{few} ($\textcolor{red}{\bullet}$) in 1\% most easily forgotten data under different training epochs.}
     \vspace{-1em}
  \label{fig:pieDistribution}
\end{figure}

\begin{figure}[t]
  \centering
   \vspace{-0.5em}
  \includegraphics[trim = 0mm 0mm 0mm 0mm, clip,scale=0.65]{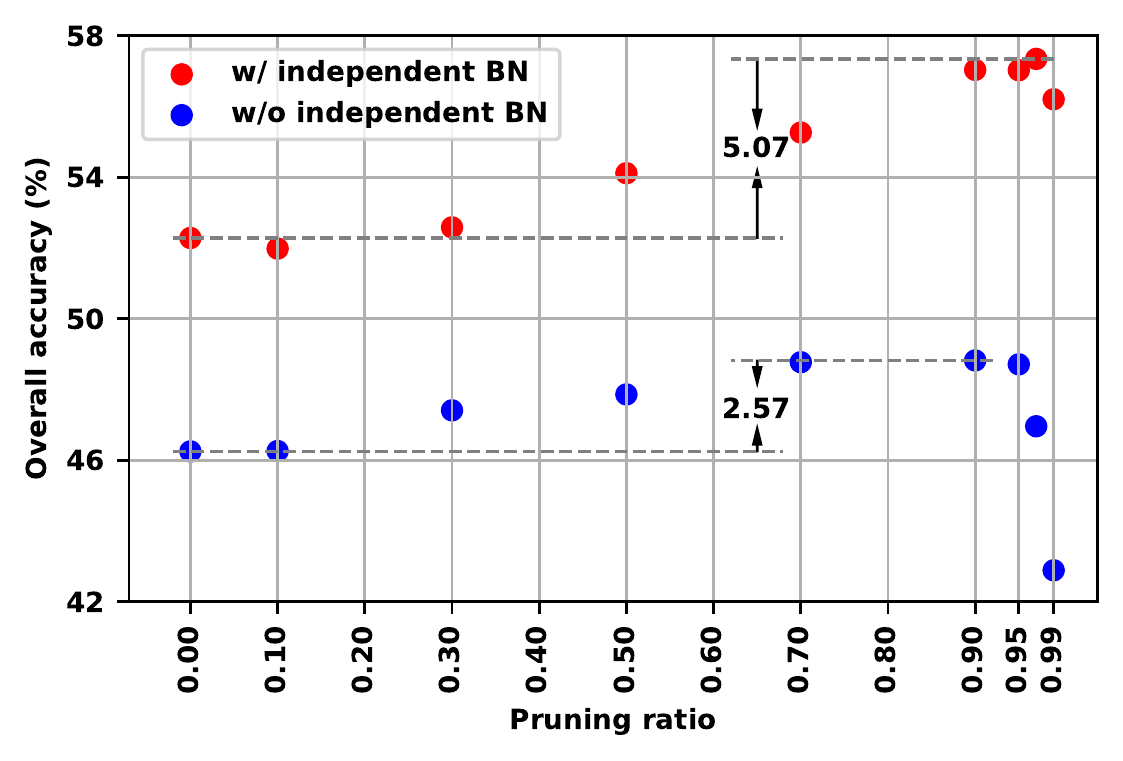}
  \vspace{-1em}
    \caption{Ablation study of \textit{linear separability performance} w.r.t. the pruning ratios for the dense branch, with ($\textcolor{red}{\bullet}$) or without ($\textcolor{blue}{\bullet}$) independent BNs per branch, on one imbalance split of CIFAR100.}
     \vspace{-0.5em}
  \label{fig:ablationAcc}
\end{figure}

We then measure the distribution of PIEs mined by the proposed SDCLR. Specifically, when pre-training on long tail splits of CIFAR100, we sample top 1\% testing data that is most easily influenced by pruning and then evaluate the percentage of \textit{many}, \textit{medium} and \textit{minor} in it under different training epochs. The difficulty of forgetting a sample is defined by the features' cosine similarity before and after pruning. Figure \ref{fig:pieDistribution} shows the \textit{minor} and \textit{medium} are much more likely to be impacted comparing to \textit{many}. In particular, while the group distributions of the found PIEs show some variations along with training epochs, in general, we find samples from the \textit{minor} group to gradually increase, while the \textit{many} group samples continue to stay low percentage especially when it is close to convergence.
%indicating the PIEs sampled by SDCLR are mainly in the tail classes.

\subsection{Sanity Check with More Baselines}

\textbf{Random dropout baseline:} To  verify whether pruning is necessary, we compare with using random dropout~\cite{srivastava2014dropout} to generate the sparse branch. Under dropout ratio of 0.9,  [\textit{linear separability}, \textit{few-shot accuracy}] are [21.99$\pm$0.35\%, 15.48$\pm$0.42\%], which are much worse than both SimCLR and SDCLR reported in Tab.~\ref{tab:pruneImproveImbalance} and~\ref{tab:pruneImproveImbalanceFewshot}. In fact, the dropout baseline is often hard to converge.

\textbf{Focal loss baseline:} We also compare with the popular focal loss~\cite{lin2017focal} for conducting this suggested sanity check. With the best grid searched gamma of 2.0 ( grid is [0.5, 1.0, 2.0, 3.0]), it decreases the [accuracy,std] of \textit{linear separability} from [47.33$\pm$0.33\%, 2.70$\pm$1.25\%] to [46.48$\pm$0.51\%, 2.99$\pm$1.01\%], respectively. Further analysis shows the contrastive loss scale is not tightly connected with the major or minor class membership as we hypothesized. A possible reason is that the randomness of SimCLR augmentations also notably affects the loss scale.

\textbf{Extending to Moco pre-training:} We try MocoV2~\cite{he2020momentum,chen2020improved} on CIFAR100-LT. The [accuracy,std] of \textit{linear separability} is [48.23$\pm$0.20\%, 3.50$\pm$0.98\%] and [accuracy,std] of \textit{few-shot performance} is [24.68$\pm$0.36\%, 6.67$\pm$1.45\%], respectively, which is worse than SDCLR in Tab~\ref{tab:pruneImproveImbalance} and~\ref{tab:pruneImproveImbalanceFewshot}. %Remarkably, the balancedness of \textit{few-shot performance} is 2.69\% worse than SDCLR.

\subsection{Ablation Studies on the Sparse Branch}
\label{ablationExps}

We study the \textit{linear separability performance} under different pruning ratios in one imbalance subset of CIFAR100. As shown in Figure.~\ref{fig:ablationAcc}, the overall accuracies consistently increase with the pruning ratio until it exceeds 90\%, which will lead to a quick drop. That shows a trade-off for the sparse branch between being stronger (i.e., needing larger capacity) and being effective in spotting more difficult examples (i.e., needing being sparse).

%no matter if BN is independent or not. This proves the pruned branch can consistently boost the performance under a reasonable range. However, if the pruning ratio exceeds a point, 0.95 for pruning w/o independent BN, the performance would sharply drop as current branch is too weak to contrast with the dense branch. When employing independent BN, the dropping point would be postponed to 95$\%$ and pruning ratio of 99$\%$ can even yield a higher performance than no pruning.

% Moreover, employing independent BN can consistently yield higher performance. It is also worth noting that the maximum improvement brought by pruning can be higher (5.07\% v.s. 2.57\%) with employing independent BN. Surprisingly, when the pruning ratio is 0, independent BN along can still bring a significant improvement, this could probably because the independent BN reduces the complexity of the optimization.

\begin{figure}[!htb]
\centering
\vspace{-0.5em}
\subfigure[]{\includegraphics[width=0.49\linewidth]{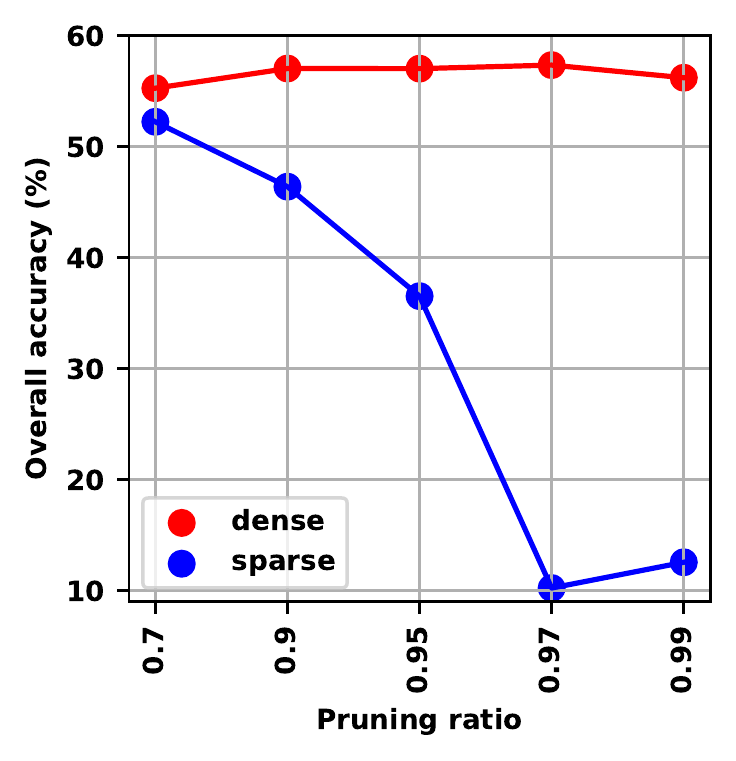}}
\hfill
\subfigure[]{\includegraphics[width=0.49\linewidth]{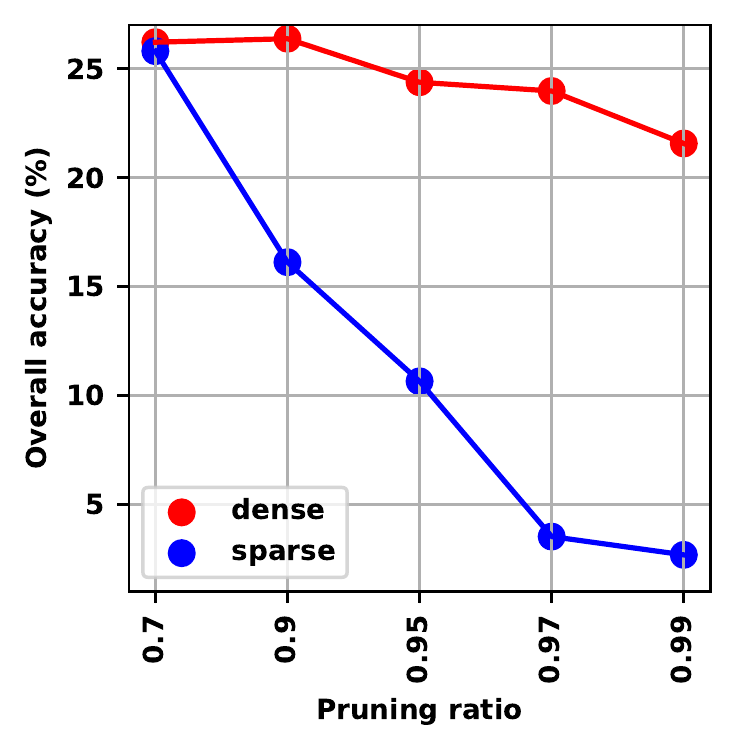}}
\vspace{-1em}
\caption{Compare (a) \textit{linear separability performance}, and  (b) \textit{few-shot performance}, for representations learned by dense ($\textcolor{blue}{\bullet}$) and sparse ($\textcolor{red}{\bullet}$) branches. Both are pre-trained and evaluated on one long tail split of CIFAR100, under different pruning ratios.}
\vspace{-0.5em}
\label{fig:ablationPrune}
\end{figure}

We also explore the \textit{linear separability and few-shot performance} of the sparse branch in Figure \ref{fig:ablationPrune}. In the linear separability case (a), the sparse branch quickly lags behind the dense branch when the sparsity goes above 70$\%$, due to limited capacity. Interestingly, even a ``weak" sparse branch can still assist the learning of its dense branch. The few shot performance also shows the similar trend.
% However, in the few-shot case (b), we can see the fine-tuning performance of the sparse and the dense branches are very close, even at very highs pruning ratios of $70\%$ to $90\%$. Note that whenever training the sparse branch, we will not recover the zero weight being pruned. Therefore, it shows sparsity itself could have a favorable regularization effect here. The sparse branch obtained by our model can be considered like a ``lottery ticket'' at the pre-trained initialization \cite{chen2020lottery}, for the downstream few-shot learning task. 

%However, when the pruning ration continue to increase, the performance of the pruned branch would be worse than its dense branch.

% show the pruned branch can also achieve good performance.
%\subsection{\Ziyu{Understanding SDCLR}}

\begin{figure}
% \vspace{-0.5cm}
\includegraphics[width=\linewidth]{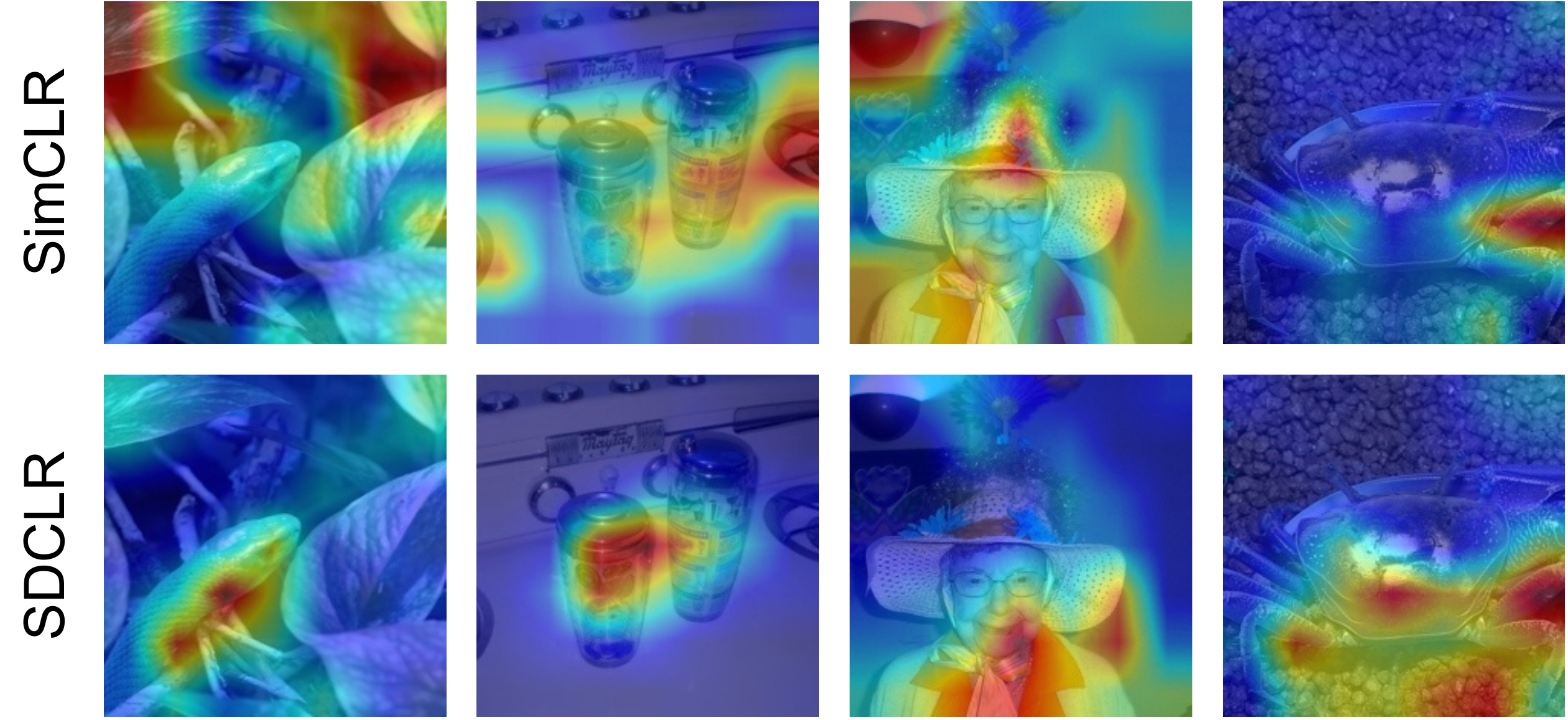}
\vspace{-0.3cm}
\caption{Visualization of attention on tail class images with Grad-CAM~\cite{selvaraju2017grad}. The first and second row corresponds to SimCLR and SDCLR, respectively.}
\label{fig:CAM}
% \vspace{-0.5cm}
\end{figure} 

\begin{figure}
% \vspace{-0.5cm}
\includegraphics[width=\linewidth]{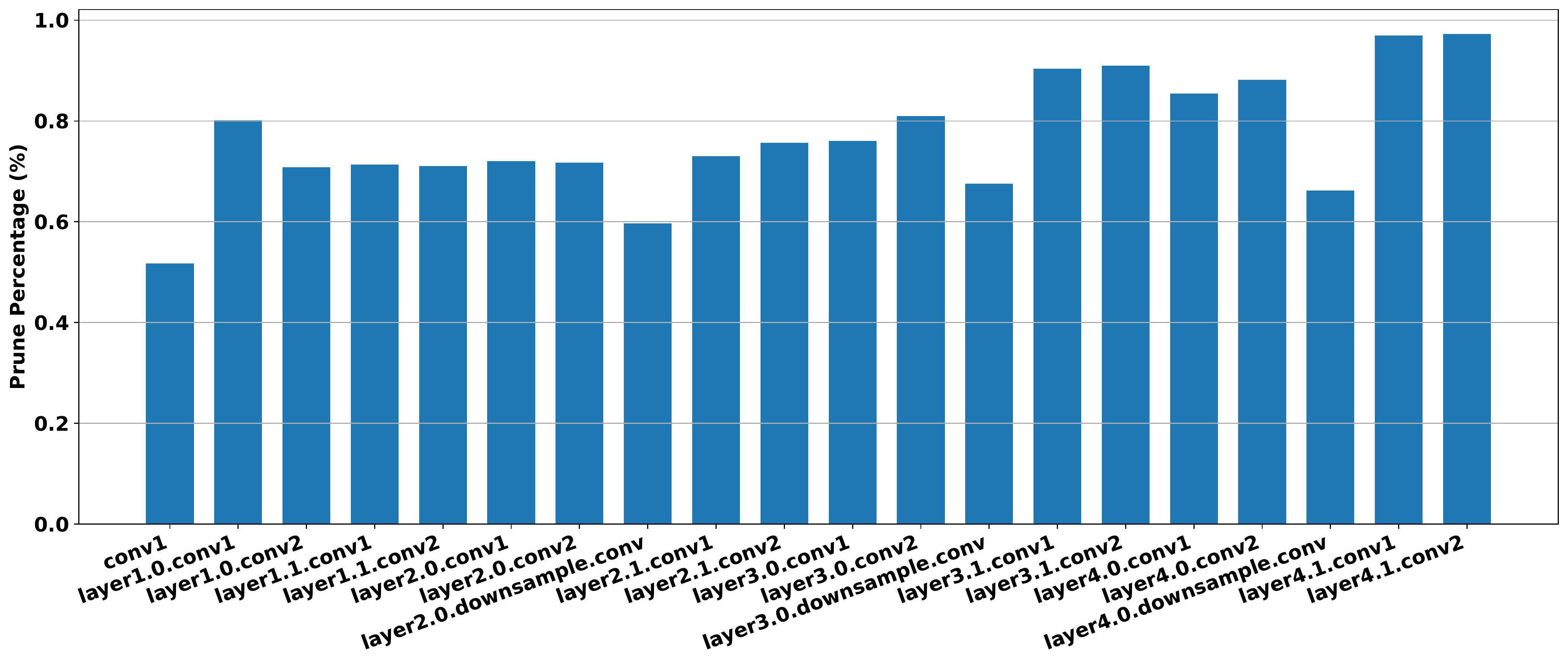}
\vspace{-0.3cm}
\caption{Layer-wise pruning ratio for SDCLR with 90\% pruning ratio on Cifar100-LT. The layer following the feed-forward order. We follow~\cite{he2016deep} for naming each layer.}
\label{fig:pruneRatios}
% \vspace{-0.5cm}
\end{figure}

\textbf{The Sparse Branch Architecture} The visualization of pruned ratio for each layer is illustrated in Figure~\ref{fig:pruneRatios}. Overall, we find the sparse branch's deeper layers to be more heavily pruned. This is aligned with the intuition that higher-level features are more class-specific. 

\textbf{Visualization for SDCLR} We visualize the features of SDCLR and SimCLR on minor classes with Grad-CAM~\cite{selvaraju2017grad} in the Figure~\ref{fig:CAM}. SDCLR shows to better localize class-discriminative regions for tail samples. 

\section{Conclusion}
In this work, we improve the robustness of Contrastive Learning towards imbalance unlabeled data with the principle framework of SDCLR. Our method is motivated the the recent findings that deep models will tend to forget the samples in the long-tail when being pruned. Through extensive experiments across multiple datasets and imbalance settings , we show that SDCLR can significantly mitigate the imbalanceness. Our future work would explore extending SDCLR to more contrastive learning frameworks.

\bibliography{example_paper}
\bibliographystyle{icml2021}

\newpage
% \documentclass{article}

% % Recommended, but optional, packages for figures and better typesetting:
% \usepackage{microtype}
% \usepackage{graphicx}
% \usepackage{subfigure}
% \usepackage[dvipsnames]{xcolor}
% % \usepackage{subcaption}
% \usepackage{booktabs, tabularx, wrapfig, adjustbox, multirow} % for professional tables
% \usepackage{amssymb}

% \newcolumntype{L}[1]{>{\raggedright\arraybackslash}p{#1}}
% \newcolumntype{C}[1]{>{\centering\arraybackslash}p{#1}}
% \newcolumntype{R}[1]{>{\raggedleft\arraybackslash}p{#1}}

% % hyperref makes hyperlinks in the resulting PDF.
% % If your build breaks (sometimes temporarily if a hyperlink spans a page)
% % please comment out the following usepackage line and replace
% % \usepackage{icml2021} with \usepackage[nohyperref]{icml2021} above.
% \usepackage{hyperref}

% % Attempt to make hyperref and algorithmic work together better:
% \newcommand{\theHalgorithm}{\arabic{algorithm}}

% % Use the following line for the initial blind version submitted for review:
% \usepackage{icml2021}
% \usepackage{enumitem}

% If accepted, instead use the following line for the camera-ready submission:
%\usepackage[accepted]{icml2021}

% The \icmltitle you define below is probably too long as a header.
% Therefore, a short form for the running title is supplied here:
% \section*{Self-Damaging Contrastive Learning}

% \section*{document}

% \newcommand\Tianlong[1]{\textcolor{blue}{[TL: #1]}}
% \newcommand\Ziyu[1]{\textcolor{red}{[#1]}}

\twocolumn[
\section*{Supplementary Material: Self-Damaging Contrastive Learning}

\vskip 0.3in
]



This supplement contains the following details that we could not include in the main paper due to space restrictions.
\begin{itemize}
\item {\bf (Sec.~\ref{sec:computingArch})} Details of the computing infrastructure.
\item {\bf (Sec.~\ref{sec:dataset})}  Details of the employed datasets.
\item {\bf (Sec.~\ref{sec:hyperparameter})}  Details of the employed hyperparameters.
\end{itemize}

\section{Details of computing infrastructure}
\label{sec:computingArch}
Our codes are based on Pytorch \cite{paszke2017automatic}, and all models are trained with GeForce RTX 2080 Ti and NVIDIA Quadro RTX 8000.

\section{Details of employed datasets}
\label{sec:dataset}
\subsection{Downloading link for employed dataset}

The datasets we employed are CIFAR10/100, and ImageNet. Their downloading links can be found in Table.~\ref{tab:download_link}.

\begin{table}[h]
\centering
\caption{Dataset downloading links}
\vspace{3mm}
    \label{tab:download_link}
    \begin{adjustbox}{max width=\linewidth}
    \begin{tabular}{@{}L{1.5cm}C{8cm}@{}}
    \toprule
    Dataset       & Link       \\ \midrule
    ImageNet      & http://image-net.org/download \\  
    CIFAR10       & https://www.cs.toronto.edu/~kriz/cifar-10-python.tar.gz  \\ 
    CIFAR100      & https://www.cs.toronto.edu/~kriz/cifar-100-python.tar.gz \\
    \bottomrule
    \end{tabular}
    \end{adjustbox}
 %   \vspace{-1em}
\end{table}

\subsection{Split train, validation and test subset}
For CIFAR10, CIFAR100, ImageNet, and ImageNet-100, the testing dataset is set as its official validation datasets. We also randomly select [10000, 20000, 2000] samples from the official training datasets of [CIFAR10/CIFAR100, ImageNet, ImageNet-100] as validation datasets, respectively.

\section{Details of hyper-parameter settings}
\label{sec:hyperparameter}

\subsection{Pre-training}

We identically follow SimCLR \cite{chen2020simple} for pre-training settings except the epochs number. On the full dataset of CIFAR10/CIFAR100, we pre-train for 1000 epochs. In contrast, on sub-sampled CIFAR10/CIFAR100, we would enlarge the pre-training epochs number to 2000 given the dataset size is small. Moreover, the pre-training epochs of ImageNet-LT-exp/ImageNet-100-LT is set as 500.

\subsection{Fine-tuning}

% \textbf{Network component choose:} When evaluating the \textit{linear separability performance}, the backbone of the network would be fixed, and we only optimize the linear head. For evaluating the \textit{few-shot performance}, we by default fine-tuning the whole network. However, for subsets of ImageNet-100, we would fine-tune the linear layer while fixing the backbone as we empirically find it would yield higher performance.

We employ SGD with momentum 0.9 as the optimizer for all fine-tuning. We follow~\cite{chen2020improved} employing learning rate of 30 and remove the weight decay for all fine-tuning.
When fine-tuning for \textit{linear separability performance}, we train for 30 epochs and decrease the learning rate by 10 times at epochs 10 and 20 as we find more epochs could lead to over-fitting. However, when fine-tuning for \textit{few-shot performance}, we would train for 100 epochs and decrease the learning rate at epoch 40 and 60, given the training set is far smaller.

\end{document}